\documentclass[letterpaper]{article} 
\usepackage{aaai25}  
\usepackage{times}  
\usepackage{helvet}  
\usepackage{courier}  
\usepackage[hyphens]{url}  
\usepackage{graphicx} 
\usepackage{natbib}  
\usepackage{caption} 
\usepackage{algorithm}
\usepackage{algorithmic}

\usepackage{tcolorbox}

\usepackage{amsmath}
\usepackage{amsfonts}
\usepackage{amsthm}
\usepackage{amssymb}
\usepackage{mathtools}
\usepackage{braket}

\theoremstyle{plain}

\newtheorem{lemma}{Lemma}

\theoremstyle{definition}

\newtheorem{assumption}{Assumption}

\theoremstyle{remark}

\newtheorem*{theorem*}{Theorem}
\newtheorem*{lemma*}{Lemma}
\newtheorem*{definition*}{Definition}
\newtheorem*{corollary*}{Corollary}
\newtheorem*{remark*}{Remark}

\DeclareMathOperator*{\E}{\mathbb{E}}

\newcommand{\s}{\mathcal{S}}
\newcommand{\R}{\mathbb{R}}
\newcommand{\A}{\mathcal{A}}

\usepackage{color}
\usepackage{multicol}
\newcommand{\JA}[1]{{\color{blue}#1}}

\newtcolorbox{theoremBox}[1][]{colback=blue!5!white, colframe=gray!50!black, 
coltitle=blue!40!black, fonttitle=\bfseries, title=#1, sharp corners, boxrule=0.8pt, 
before=\par\smallskip\noindent, after=\par\smallskip}

\usepackage{newfloat}
\usepackage{listings}
\DeclareCaptionStyle{ruled}{labelfont=normalfont,labelsep=colon,strut=off} 
\floatstyle{ruled}
\newfloat{listing}{tb}{lst}{}
\floatname{listing}{Listing}
\title{EVAL: EigenVector-based Average-reward Learning}

\author{
    Jacob Adamczyk\textsuperscript{\rm 1,2}, Volodymyr Makarenko\textsuperscript{\rm 3}, Stas Tiomkin\textsuperscript{\rm 4}, Rahul V. Kulkarni\textsuperscript{\rm 1,2}
}
\affiliations{
    \textsuperscript{\rm 1}Department of Physics, University of Massachusetts Boston\\
    \textsuperscript{\rm 2}The NSF Institute for Artificial Intelligence and Fundamental Interactions\\
    \textsuperscript{\rm 3}Department of Computer Engineering, San Jos\'e State University\\
    \textsuperscript{\rm 4}Department of Computer Science, Whitacre College of Engineering, Texas Tech University

    jacob.adamczyk001@umb.edu, volodymyr.makarenko@sjsu.edu, stas.tiomkin@ttu.edu, rahul.kulkarni@umb.edu
}

\begin{document}

\maketitle

\begin{abstract}

In reinforcement learning, two objective functions have been developed extensively in the literature:  discounted and averaged rewards. The generalization to an entropy-regularized setting has led to improved robustness and exploration for both of these objectives. Recently, the entropy-regularized average-reward problem was addressed using tools from large deviation theory in the tabular setting. This method has the advantage of linearity, providing access to both the optimal policy and average reward-rate through properties of a single matrix. In this paper, we extend that framework to more general settings by developing approaches based on function approximation by neural networks. This formulation reveals new theoretical insights into the relationship between different objectives used in RL. Additionally, we combine our algorithm with a posterior policy iteration scheme, showing how our approach can also solve the average-reward RL problem without entropy-regularization. Using classic control benchmarks, we experimentally find that our method compares favorably with other algorithms in terms of stability and rate of convergence.






\end{abstract}





         






\maketitle 
\section{Introduction}

To solve the central problem of reinforcement learning, an agent continuously and autonomously interacts with its surroundings to maximize a long-term reward signal. The standard method of solving reinforcement learning (RL) tasks involves a discounted objective function: the agent seeks to maximize an infinite discounted sum of rewards, as expected under a chosen control policy.
This geometrically discounted sum ensures a convergent objective function, making it a convenient representation of the problem to be solved. The discounted RL literature is vast in both theoretical \cite{sutton&barto, kakade2003sample, bertsekas2012dynamic} and algorithmic \cite{schulman2015high, andrychowicz2020matters} studies and has demonstrated great success in real-world problems of interest \cite{nature-dqn, trpo, ppo, rainbow, SAC1}. However, in many RL problems, this use of discounting is simply a useful proxy for the solving the true objective function: maximization of total episode reward. As a result, the actual choice of the discount factor, $\gamma$, is unphysical, not grounded in any corresponding physical timescale. As such, it is treated as a hyperparameter which is often tuned over for best performance (or set to some fixed value, e.g. $\gamma=0.99$, for simplicity). There is a precedent in past work for rejecting the discounted framework \cite{naik2019discounted}, for example the Heaven and Hell MDP~\cite{schwartz1993reinforcement}, suggesting that in many tasks the use of discounting is not only unphysical, but can lead to disastrous outcomes. Thus in long-term (``continuing'') decision-making settings, another objective is needed.

An alternative approach to ensuring convergence of rewards across infinitely-long trajectories is to instead use the \textbf{average-reward} objective function~\cite{mahadevan1996average}. {Despite offering a principled alternative to the long-term optimization problem, the average reward framework has not been historically popular in the RL literature, perhaps due to the lack of successful algorithms in this setting}.
However, recent work has developed new algorithms and theory \citep{ARTRPO, APO, ARO-DDPG, PRR, wan2021learning, naik2024reward} specifically for the average-reward objective. This prior work has focused on the tabular setting or on using policy-gradient techniques to develop new algorithms. Although these methods have proven useful in their respective domains, there remain gaps in the field that can be addressed with new approaches.

One such gap in the field of average-reward algorithms is the lack of entropy-regularized objectives, which have become a cornerstone of state-of-the-art algorithms in discounted RL~\cite{HaarnojaSQL, SAC1, ppo, trpo}. Formally, entropy-regularization involves including an ``information'' cost for deviating from a pre-specified (e.g. prior, guide, or behavioral) policy. This entropy cost is weighted with an inverse temperature parameter, $\beta$. This entropy-based regularization, in combination with treating the rewards as negative energies, reveals a close connection to statistical mechanics~\cite{PRR, rose2021reinforcement}. Furthermore, including an entropy regularization term in the RL objective leads to more robust solutions (non-greedy optimal policies) and has theoretically been shown to be more adaptable to changes in reward and transition dynamics~\cite{eysenbach2022maximum}. Preventing the optimal policy from collapsing to a deterministic function can ensure additional exploration occurs during training and deployment, often leading to faster learning~\cite{ahmed2019understanding}.




\begin{figure}[t]
    \centering
    \includegraphics[width=0.45\textwidth]{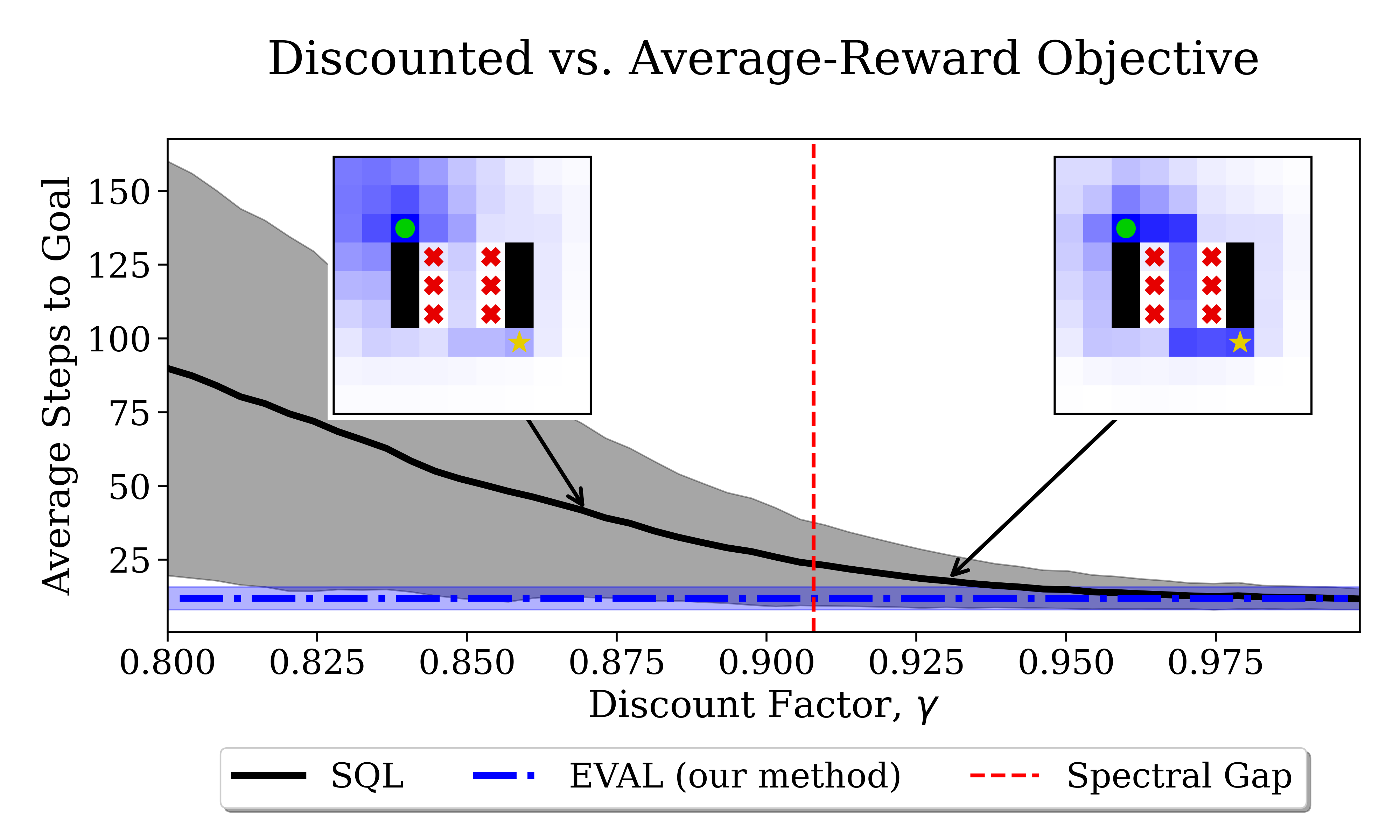}
        \caption{Performance of discounted soft Q-Learning (SQL) as a function of discount factor compared with solution using the proposed average-reward method (EVAL). Note that the average-reward solution (blue line) recovers the discounted solution as $\gamma~\to~1$. For the discounted objective, computational cost grows as $(1~-~\gamma)~^{-1}$ and choosing a low discount factor to reduce computational cost can result in lower rewards. The boundary of low reward, low complexity and high reward, high complexity is demarcated by the discount factor derived from the spectral gap of the associated tilted matrix (cf. ``Preliminaries''). Insets: state-occupation distributions following the SQL optimal policies at $\gamma=0.87, 0.93$. Green dot denotes initial position of agent, and star denotes the goal. The agent can move in any of the cardinal directions. Since we use entropy-regularization, the optimal policy is stochastic, thus yielding a variance in the return (plotted with a shaded interval for each method). Inverse temperature $\beta=15$.}
    \label{fig:disc-vs-avg_reward}
\end{figure}


In this paper, we introduce a novel algorithm for average-reward RL with entropy regularization:
EigenVector-based Average-reward Learning (EVAL).
Figure~\ref{fig:disc-vs-avg_reward} illustrates a simple experiment highlighting one benefit of the EVAL approach to entropy-regularized RL compared to the discounted approach \cite{haarnoja2017reinforcement}. When the RL problem is posed in the discounted framework, a discount factor $\gamma$ is a required input parameter. However, there is often no principled approach for choosing the value of $\gamma$ corresponding to the specific problem being addressed. Thus, the experimenter must treat $\gamma$ as a hyperparameter. This reduces the choice of $\gamma$ to a trade-off between large values to capture long-term rewards and small values to capture computational efficiency which typically scales polynomially with the horizon, $(1-\gamma)^{-1}$ \cite{kakade2003sample}. The horizon introduces a natural timescale to the problem, but this timescale may not be 
well-aligned with the timescale corresponding to the optimal dynamics: the mixing time of the induced Markov chain. For the discounted approach to accurately estimate the optimal policy, the discounting timescale (horizon) must be larger than the mixing time. However, estimating the mixing time for the optimal dynamics can be challenging in the general case, even when the transition dynamics are known. 
Therefore, an arbitrary ``sufficiently large'' choice of $\gamma$ is often made without knowledge of the relevant problem-dependent timescale. This can be problematic from a computational standpoint as evidenced by recent work \cite{jiang2015dependence, ppo, andrychowicz2020matters}.

The approach outlined in this work for average-reward RL (EVAL) has the added benefit that it can lead to an estimation of the mixing timescale for the optimal dynamics, which can then be used to inform the choice of a discount factor, if the discounted objective is of interest.
In Figure~\ref{fig:disc-vs-avg_reward}, the discount factor set by the spectral gap of the tilted matrix (see Eq.~\eqref{eq:tilted-matrix} for definition) is indicated by the vertical line. We empirically find that this ``spectral gap discount factor'' naturally separates discount factors between ``small'' and ``large'' values, seen by the distinct change in the optimal state distributions (inset). The discount factor set by the spectral gap indicates a point of diminishing return for further increasing $\gamma$ in soft Q-Learning (solid black line). For comparison, we plot the return given by our average-reward algorithm (dashed blue line). Importantly, we see that the average-reward solution recovers the discounted solution in the $\gamma \to 1$ limit, as expected \cite{blackwell1962discrete, mahadevan1996average}.


Despite the desirable features of both the average-reward and entropy-regularized objectives, the combination of these formulations~\cite{neu2017unified} is not as well-studied, and no function approximator algorithms exist for this setting. 
Here, we extend the ideas introduced in \cite{rawlik-thesis, argenis-thesis} to develop a new off-policy learning-based approach to find the optimal policy of average-reward MDPs. Our main contributions are as follows: 

\noindent\textbf{Main Contributions}
\begin{itemize}
    \item We provide a novel off-policy solution to average-reward RL in both entropy-regularized and un-regularized settings with general function approximators.
    \item We demonstrate experimentally the advantage of our approach against standard baselines in classic control environments.
\end{itemize}

Notably, our implementation requires minimal changes relative to the common DQN setup, making it accessible 
for researchers and allowing for multiple future extensions.

\section{Preliminaries\label{sec:preliminaries}}
Reinforcement learning treats decision-making problems under the framework of Markov Decision Processes (MDPs). In this section, we provide a brief account of the problem to be solved (the entropy-regularized average-reward) objective under this framework. As a point of notation, we will denote the set of distributions over a generic space $\mathcal{X}$ as $\Delta(\mathcal{X})$.  To begin, we describe the defining quantities of an MDP: the state space $\s$ denotes all possible configurations of the agent within its environment; the initial state distribution $\mu \in \Delta(\s)$ describes the initialization of the agent at each episode; the action space $\A$ denotes the set of allowed controls the agent may exert to affect its environment; the transition function (also ``dynamics'') is a function $p: \s \times \A \to \Delta(\s)$ describing the probability of transitioning into a future state given a current state and action; and the reward function $r: \s \times \A \to \R$ is a real-valued scalar provided to the agent at each time-step, encoding the desired behavior of the agent. 
In the following equations, we focus on the discrete state-action case for ease of notation. The applicability of the framework for continuous spaces is demonstrated empirically below.

Now, we state some of the usual assumptions for average-reward MDPs:
\begin{assumption}
The Markov chain induced by the dynamics $p$ and any stationary policy $\pi$ is irreducible and aperiodic. \label{assump:mc}
\end{assumption}

\begin{assumption}
The reward function is upper bounded.~\footnote{In the following we assume the reward is non-positive, which is equivalent to $r(s,a)$ having a finite upper-bound: the reward function can always be shifted to be entirely negative without affecting the optimal policy, merely shifting the reward rate accordingly.}
\end{assumption}

As in the discounted case, one seeks a control policy $\pi$ which maximizes the expected return. In the discounted formulation, this objective is defined as a discounted infinite sum; but in the average-reward formulation, we instead consider the limiting value of the average trajectory reward, for increasingly long trajectories:
\begin{equation}
    \sum_{t=0}^\infty \gamma^t r_t \to \lim_{N\to\infty} \frac{1}{N}\sum_{t=0}^{N-1} r_t
\end{equation}
This \textit{reward rate}, $\rho^\pi$, becomes the new objective:
\begin{equation}
    \rho^\pi=\lim_{N \to \infty} \frac{1}{N}  \E_{\tau \sim{} \pi,p,\mu} \left[\sum_{t=0}^{N-1} r(s_t, a_t) \right] = \E_{\nu} \left[ r(s,a) \right], \label{eq:reward-rate-defn}
\end{equation}
with the expectation above taken with respect to the probability of a trajectory as generated by the dynamics $p$, fixed policy $\pi$, and initial state distribution $\mu$. In the rightmost expression, $\nu$ is used to denote the stationary state-action distribution induced by the transition dynamics and choice of control policy. Because of Assumption~\ref{assump:mc}, this distribution is well-defined as the eigenvector of the stochastic transition matrix $P^\pi$:
\begin{equation}
    \sum_{s,a} p(s'|s,a) \pi(a'|s') \nu(s,a) = \nu(s',a'),
\end{equation}
or in a more compact notation, $P^\pi \nu= \nu$. Related eigenvector equations will be our primary concern in approaching the entropy-regularized average-reward objective.

With the scalar reward-rate (``bias'') defined, we now turn to the corresponding ``differential'' value function, which plays the role of the standard $Q$ function in discounted RL. Specifically, the $(s,a)$-dependent contribution to a trajectory's value is denoted as $Q_\rho^\pi$.
\begin{equation}
    Q^\pi_\rho(s,a)=\E_{\tau \sim{} p,\pi} \left[ \sum_{t=0}^\infty  r(s_t,a_t) - \rho^\pi \Biggr| s_0=s,a_0=a \right]
    \label{eq:differential-value}
\end{equation}

We now consider a more general version of this MDP which includes an entropy regularization term. Note that the original objective can be recovered in the zero temperature limit ($\beta \to \infty$). For convenience we will refer to entropy-regularized average-reward MDPs as ERAR MDPs. The ERAR MDP possesses the same components as an average-reward MDP as described, in addition to a pre-specified prior policy\footnote{We assume that $\pi_0$ has full support across $\A$, ensuring the Kullback-Liebler divergence between any policy $\pi$ and $\pi_0$ remains finite.} $\pi_0: \s \to \Delta(\A)$ and ``inverse temperature'', $\beta\in \mathbb{R}_{>0}$. The modified objective function for an ERAR MDP now includes a relative entropy based regularization term,
such that the agent now aims to optimize the expected \textit{entropy-regularized reward-rate}, denoted $\theta^\pi$ below:
\begin{equation}
    \theta^\pi=\lim_{N \to \infty} \frac{1}{N} \E_{\tau \sim{} p,\pi, 
    \mu} \left[ \sum_{t=0}^{N-1}  r(s_t,a_t) - \frac{1}{\beta} \log \frac{\pi(a_t|s_t)}{\pi_0(a_t|s_t)}  \right]. \label{eq:theta-defn}
\end{equation}
The optimal policy for an ERAR MDP is defined analogously as the solution to
\begin{equation}
    \pi^*(a|s) = \mathop{\mathrm{argmax}}_{\pi} \theta^\pi.
\end{equation}
From the above, we see that the agent must balance the returns given by directly optimizing the reward with the cost of deviating from the prior policy $\pi_0$. This also emphasizes the dependence of $\pi_0$ on the resulting optimal policy. In most work on entropy-regularized RL, the uniform distribution $\pi_0(a|s)\propto 1$ is used to express the prior belief that all actions are equally likely (preferable). However, this ``maximum entropy'' (MaxEnt) objective has a lack of flexibility, not allowing non-trivial priors or ``guide'' policies to be exploited, despite its apparent theoretical and experimental advantages ~\cite{AAAI, yan2024efficientreinforcementlearninglarge,ahmed2019understanding, grau2018soft, wu2019behavior}.  In the following, we show that the more general non-uniform prior $\pi_0$ plays a crucial role in solving the un-regularized problem.

Our Assumption~\ref{assump:mc} guarantees the expression in Equation~\eqref{eq:theta-defn} is indeed independent of the initial state-action. If on the other hand the induced Markov chain were reducible, there may be some recurrent classes whose long-term reward rate depends on the initial state and action. 

In the following, for convenience, we will simply write $\theta=\theta^{\pi^*}$ for the optimal entropy-regularized reward-rate. Now, in comparing to Equation~\eqref{eq:reward-rate-defn}, we see that the present objective, the ``ERAR rate'', includes an entropic contribution: the relative entropy between the control and prior policies: $\mathrm{KL}(\pi|\pi_0)$, where $\mathrm{KL}$ denotes the Kullback-Liebler divergence.

Corresponding to Eq.~\eqref{eq:differential-value}, the differential entropy-regularized action-value function is then given by~\footnote{we have suppressed the conditioning on the initial state and action for brevity, but the conditioning is identical to that of Eq.~\eqref{eq:differential-value}.}
\begin{equation}
    Q^\pi_\theta(s,a)=\E_{\tau \sim{} p,\pi} \left[ \sum_{t=0}^\infty  r(s_t,a_t) - \frac{1}{\beta} \log \frac{\pi(a_t|s_t)}{\pi_0(a_t|s_t)} - \theta^\pi \right] 
    \label{eq:theta-differential-value}
\end{equation}
In this section, $\theta$ and $\rho$ subscripts were used to distinguish the two (un-regularized and regularized) differential value functions. In the following sections, we omit the subscript as we focus solely on the entropy-regularized objective. Similarly, we use the shorthand $Q(s,a)=Q^{\pi^*}_\theta(s,a)$.

\section{Prior Work}
Classical approaches to the average-reward problem typically involve the following logic: Solve the discounted MDP for discount factors $\gamma_1 < \gamma_2 \cdots < 1$. As the value of the discount factor approaches $1$, the (appropriately re-scaled) optimal value function will approach that of the average-reward MDP~\cite{blackwell1962discrete}. Later work directly approaches the average-reward objective by attacking the associated Bellman equation head-on~\cite{mahadevan1996average}. Indeed, \citep{schwartz1993reinforcement} introduced \textit{R-learning} (without proof of convergence), emphasizing the benefits of the average-reward algorithm: faster reward propagation, better value disambiguation, simpler learning dynamics near initialization, and potentially speedups. \citet{schwartz1993reinforcement} also mentions how the discounted framework can be subsumed in the more general average-reward problem (Sec. 6.5 therein).

More recently, a variety of papers have considered the average-reward problem in a modern light, beginning to understand the theoretical properties of this objective and associated algorithms: \cite{even2009online, politex,abounadi,neu2017unified,wan2021learning, zhang2021average}. We note this set of prior works addresses the average-reward objective in the case of tabular settings, with discrete, finite state and action spaces. To address the continuous setting, recent work has developed \cite{ARTRPO, APO, saxena2023off} a variety of policy-based methods. In contrast, we will focus on the value-based setting here, learning the $Q$ function directly in an off-policy manner. It is worth noting that these studies have found average-reward algorithms to be superior to discounted methods in continuous control Mujoco \cite{todorov2012mujoco} tasks. 
A more complete description of the history of average-reward algorithms can be found in a recent survey: \cite{dewanto2020average}.

Alongside the average-reward literature, the discounted objective has seen the introduction of many techniques to generally improve sample complexity, by tackling planning~\cite{hafner2023mastering}, estimation bias\cite{van2016deep}, and exploration~\cite{park2023controllability}. Prior work has shown that the addition of an entropy-regularization term can improve robustness~\cite{eysenbach2022maximum}, exploration~\cite{eysenbach2018diversity}, composability~\cite{haarnoja2018composable} and sample complexity~\cite{SAC1}. The theory of entropy regularization also has a long history, discussed in a series of work such as \cite{ziebart-thesis, rawlik-thesis, todorov, todorov-pnas,SAC1,geist2019theory}. This innovation yields controllably stochastic optimal policies, a flexibility which has led Soft Actor-Critic \cite{SAC2} and its variants to become state-of-the-art solution methods for addressing the discounted objective.
To date, there are no direct combinations of the average-reward and entropy-regularized objectives for deep reinforcement learning. However, recent work by \cite{rawlik-thesis,rose2021reinforcement,PRR} has established a framework for combining these formulations. In this work, we leverage the results derived in combination with function approximation techniques to establish value-based algorithms for the average-reward objective. In the following section we give an overview of the solution which builds on results obtained in \cite{PRR} before introducing our algorithms.

\section{Solution Method\label{sec:solution}}
The presented solution method is based on insights and results from recent work \cite{mitter2000duality,rawlik2012stochastic, rose2021reinforcement, PRR,levine2018reinforcement} establishing a connection between the optimization problems in RL and statistical mechanics. Specifically, there is a mapping at the trajectory level between free energy minimization in statistical mechanics and (entropy-regularized) average reward maximization\footnote{Since energies can be viewed as negative rewards, these optimization problems are formally equivalent.}. This mapping is based on control-as-inference approaches to RL as outlined below.

The control-as-inference perspective of RL reviewed in \cite{levine2018reinforcement} introduces a binary optimality variable to solve the MaxEnt RL problem for the case of deterministic dynamics. Specifically, Bayesian inference of the posterior dynamics resulting from conditioning on this optimality variable gives the solution to the optimal control problem. Furthermore, this approach also reveals that the optimal trajectory distribution follows a Boltzmann (softmax) distribution, similar to trajectory distributions arising from generalizations of the canonical ensemble in statistical mechanics \cite{rose2021reinforcement, chetrite2015variational, chetrite2015nonequilibrium}.
 
This connection to statistical mechanics indicates that the optimal trajectory distribution can be generated by a ``tilted'' transition matrix, which encodes information about the RL solution, as shown in \cite{PRR}. Specifically, the trajectory distribution corresponding to the optimal policy for the ERAR objective can be generated from the tilted transition matrix:

\begin{equation}
    \widetilde{P}_{(s',a'),(s,a)} = p(s'|s,a) \pi_0(a'|s') e^{\beta r(s,a)}.\label{eq:tilted-matrix}
\end{equation} 
 
For a fixed value of $\beta$, the tilted matrix generates the dynamics for trajectories having an average return that is ``tilted'' or biased away from their typical value. Thus, by tuning $\beta$, the matrix $\widetilde{P}$ gives access to those trajectories with significantly larger expected returns than those obtained by running the system's unbiased dynamics: $p(s'|s,a) \pi_0(a'|s')$.

Denoting the dominant eigenvalue and corresponding eigenvector of $\widetilde{P}$ as $e^{\beta \theta}$ and $u(s,a)$, the solution to the ERAR MDP is given by:

\begin{lemma}[\citet{PRR}]
The entropy-regularized average reward rate is given by $\theta$, and the optimal differential value function and optimal policy are given by:
\begin{align}
    Q(s,a) &= \frac{1}{\beta} \log u(s,a)\; , \label{eq:optimal-q} \\
    \pi(a|s) &= \frac{\pi_0(a|s) u(s,a)}{\sum_a \pi_0(a|s) u(s,a)}. \label{eq:optimal-policy}
\end{align}
\label{lem:prr}
\end{lemma}

The solution method used requires the transition dynamics to be deterministic, though we discuss the stochastic generalization at the end of this section.
It should also be noted that this method obtains the differential value and corresponding policy for the \textit{optimal} entropy-regularized reward-rate, as opposed to a policy-evaluation technique, which may gather information on arbitrary suboptimal policies. 

To gain further insights into the utility of the tilted matrix, consider the class of allowed trajectories (of length $N$) that start at a given state-action pair $(s,a)$ and terminate at $(s',a')$. If we condition on optimality throughout the trajectory, then the posterior probability of observing this class of trajectories can be shown~\cite{PRR} to be proportional to $\left[\widetilde{P}^N \right]_{(s',a'),(s,a)}$. In the long-time limit (Eq.~\eqref{eq:theta-defn}), the probability for this class of trajectories is well-approximated by the dominant contribution in the spectral decomposition (thus corresponding to the largest eigenvalue) of $\widetilde{P}$:
\begin{equation}
    \left[\widetilde{P}^N \right]_{(s',a'),(s,a)} \approx e^{N\beta \theta} u(s,a) v(s',a'),
    \label{eq:ptilda-largeN}
\end{equation} where the left and right dominant eigenvectors $u,v$ correspond to the dominant eigenvalue (Perron root) $\exp{\left(\beta \theta\right)}$. Now, to connect to the solution of the entropy-regularized RL problem, we recall that the soft value function is given by the log-probability of a trajectory being optimal~\cite{levine2018reinforcement}, given the initial state-action $(s,a)$. To obtain the corresponding probability, we thus need to sum over all possible final state-action pairs $(s',a')$ in Eq.~\eqref{eq:ptilda-largeN}. Taking the logarithm then gives the soft value function as shown in Eq. (21) of \cite{PRR}. This leads to the identification of $\theta$ as the ERAR rate and the derived results shown in Lemma~\ref{lem:prr}. 

\nocite{rose2021reinforcement}
 


In practice it may not be possible to construct this tilted matrix beyond the tabular setting without model-based techniques. Nevertheless, only a single component of its spectrum is required for an analytic solution to the ERAR MDP. As such, we will consider only the relevant component of the spectrum to devise a model-free learning algorithm, circumventing the need for learning the entire matrix.

The preceding discussion shows that the solution to the ERAR MDP is given by solving an eigenvalue equation for the tilted dynamics:

\begin{equation}
    e^{\beta \theta} u(s,a) = \sum_{s',a'} u(s',a') \widetilde{P}_{(s',a'),(s,a)}. \label{eq:eigvec-tilted}
\end{equation}
Because of the assumptions imposed, the Perron-Frobenius theorem applies to $\widetilde{P}$, guaranteeing a \textit{unique} positive left eigenvector $u$ and well-defined ERAR rate $\theta$.
We can rewrite the above equation in the following form, reminiscent of a temporal-difference backup equation:
\begin{equation}
    u(s,a) = e^{\beta(r(s,a) - \theta)}\E_{s'\sim{}p, a'\sim{} \pi_0} u(s',a') ,
    \label{eq:eigvec}
\end{equation}
for an eigenvector $u > 0$ and (the unique real) eigenvalue with $\theta < 0$. The next largest eigenvalue (in magnitude), denoted $e^{\beta \xi}$, determines the spectral gap of $\widetilde{P}$: $(\beta \xi - \beta \theta)^{-1}$. As discussed in the Introduction, the spectral gap determines an important timescale for the optimal dynamics. This mixing time controls the rate of convergence to the ``(quasi) steady-state distribution'' ($v$ above) \cite{meleard2012quasi, PRR} and hence controls the effective time horizon of the agent in its environment.

Given that we do not rely on policy-gradient techniques, and since the expectation in Equation~\eqref{eq:eigvec} is over the pre-specified prior policy $\pi_0$, we have an off-policy algorithm whose trajectory data can be collected by any rollout policy. Note that in log-space, this equation resembles the TD equation for soft Q-learning, without a discount factor, and with an additional correction for the reward-rate, $\theta$.

As discussed by \cite{wan2021learning}, all average reward algorithms (except their Centered Differential TD-learning) are only able to learn an ``uncentered'' value function: i.e. a differential value function that may be off by some global constant. In contrast, since our solution method is based on a dominant eigenvalue and corresponding normalized eigenvector (guaranteed to exist through the Perron-Frobenius theorem, and normalized as a probability against the corresponding right eigenvector), our algorithm necessarily finds a ``centered'' differential value function for ERAR MDPs.

For the case of stochastic transition dynamics, the policy derived from the tilted matrix is optimal only if the agent can also change the transition dynamics correspondingly. However, controlling the dynamics is often infeasible and the transition dynamics is generally assumed to be fixed by the problem statement. 
Through Bayesian inference and matching the two objective functions, \cite{UAI} shows how to address this issue by iteratively biasing the reward function and transition dynamics of the original MDP so that the solution for a controlled problem coincides with the uncontrolled transition dynamics. To avoid these complications, in the following we focus on the case of deterministic transition functions. We discuss possible extensions to stochastic dynamics for future work in the penultimate section.

\section{Proposed Algorithms\label{sec:algos}}

In this section, we present pseudocode of our proposed algorithms. We would like to highlight that the core of the algorithms here are built on DQN \cite{nature-dqn}. This means that existing codebases can easily adapt DQN-style methods with significant advantage (Fig.~\ref{fig:classic-control}) with limited additional complexity. However, there are several important distinctions specific to the ERAR objective. In the next section we present Algorithm \ref{alg:u} which highlights these differences in {\color{red}red}. In the following section, we present an algorithm to solve the un-regularized average-reward objective. In the Appendix, the pseudocode for Algorithm~\ref{alg:u-ppi} highlights the differences from Algorithm~\ref{alg:u} in {\color{blue}blue}.

\subsection{Solution to ERAR-MDP\label{sec:erar-algo}}
Our first algorithm implements three key components present in many value-based deep RL methods: (1) an estimate of the value function (left eigenvector) parameterized by two deep neural nets (inspired by \cite{van2016deep}), (2) stochastic gradient descent on a temporal-difference error with Adam \cite{KingBa15} and (3) a replay buffer of stored experience. In principle, the replay buffer can be collected by any behavior policy such as an $\varepsilon$-greedy policy \cite{nature-dqn}, but we use the learnt policy (Eq.~\eqref{eq:optimal-policy}) for more effective exploration as in \cite{SAC1}. Since we use an average-reward objective, we must also maintain a running estimate of the entropy-regularized reward-rate, $\theta$.
\begin{algorithm}[t]
  \caption{EVAL}
  \label{alg:u}
  \begin{algorithmic}[1]
    \STATE \textbf{IN:} sample budget, environment, $\beta$, hyperparameters
    \STATE \textbf{Initialize}:
        \STATE Online network weights: $\psi_i \sim $ init. distribution
        \STATE Target network weights: $\bar{\psi}_i = \psi_i$
        \STATE Entropy-regularized reward-rate: $\theta = 0$
        \STATE Replay buffer: $\mathcal{D} = \{\}$
        
    \WHILE {$t < $ sample budget}
        \STATE Collect experience:
            \STATE Sample action $a \sim \pi_0$
            \STATE Take step in environment $s' \sim p(\cdot | s,a)$
            \STATE Save to replay buffer: $\mathcal{D} \leftarrow \{s, a, r, s'\}$
            \IF{train this step}
            \FOR{each gradient step}
                \STATE Sample minibatch  $\mathcal{B} \subset \mathcal{D}$ 
                \STATE {\color{red}Calculate loss via Eq.~\eqref{eq:u-loss}}
                \STATE Do gradient descent on $u$ network(s)
                \STATE {\color{red}Calculate $\theta_\text{new}$ via  Equation~\eqref{eq:theta-batch} }
                
            \ENDFOR
        \ENDIF
        {\color{red}\IF{update $\theta$}
        \STATE Update $\theta$ estimate: $\theta \gets \theta \cdot(1-\tau_\theta) + \theta_{\text{new}}\cdot \tau_\theta$
        \ENDIF}
        \IF{update target parameters}
        \STATE Update target parameters (Polyak averaging with parameters $\tau_\psi$).
        \ENDIF
    \ENDWHILE
    \STATE \textbf{OUT:} Optimal policy for ERAR-MDP
  \end{algorithmic}
\end{algorithm}

Our model-free algorithm uses the update equations prescribed by \cite{PRR} (cf. Appendix~\ref{app:theory}) to learn $u(s,a)$ and $\theta$ through stochastic approximation.  

For updating the $u$ network, we use a mean squared error loss between the online network and corresponding estimate calculated through a target network:
\begin{equation}\label{eq:u-loss}
    \mathcal{J}(\psi) = 
    \frac{1}{2}\E_{s,a \sim{} \mathcal{D}} \left(u_\psi(s,a) - \hat{u}_{\bar{\psi}}(s,a) \right)^2.
\end{equation}
We denote the trained online network as $u_\psi$ and its temporal difference (TD) target as $\hat{u}_{\bar{\psi}}$ (we use $\hat{u}$ to emphasize this is not a neural net itself, but a stand-in for the target value calculated by Eq.~\eqref{eq:u-update}). The TD target is calculated based on the lagging network's weights, denoted $\bar{\psi}$ (cf. Appendix~\ref{app:experiments} for further implementation details). To find the TD target equation, we read off Equation~\eqref{eq:eigvec} as:
\begin{equation}
     \hat{u}_{\bar{\psi}}(s,a) = e^{\beta \left( r(s,a) - \theta \right)} \E_{s' \sim{} p, a' \sim{} \pi_0} u_{\bar{\psi}}(s',a').
     \label{eq:u-update}
\end{equation}

The value of $\theta$ must be updated online as well. To this end, we re-interpret Eq.~\eqref{eq:eigvec} as an equation for the entropy-regularized reward-rate, $\theta$. Since such an equation is valid for any $s,a$, we can use the entire batch of data $\mathcal{B}$ (sampled uniformly from the replay buffer) to obtain a more accurate estimate for $\theta$. This $(s,a)$-dependent calculation of $\theta$ is based on the current estimate of the left eigenvector, $u_\psi$ (online network as opposed to target network). To preserve the linear structure of the eigenvector equation (Eq.~\eqref{eq:eigvec-tilted}), we propose to track $\theta$ through the eigenvalue itself (exponential~of~$\theta$): 
\begin{equation}
    e^{\beta\theta} = \frac{1}{|\mathcal{B}|}\sum_{\{s,a,r,s'\}\in \mathcal{B}} \frac{e^{\beta r}\E_{a' \sim{} \pi_0 } u_\psi(s',a')}{u_\psi(s,a)}\; .
    \label{eq:theta-batch}
\end{equation}

The value of $\theta$ is updated after averaging its value over the ``gradient steps'' loop in Algorithm \ref{alg:u}.
To improve stability of the next iteration of Equation~\eqref{eq:u-update}, we find it helpful to allow $\theta$ to slowly mix with previous estimates, and hence use a (constant) step size to update its value (Line 21 in Algorithm~\ref{alg:u}).

Inspired by \cite{van2016deep, fujimoto2018addressing}, we will train two online networks in parallel. 
We find this to help considerably improve the evaluation reward, as shown in Figure~\ref{fig:ablation-num-nets}.  Interestingly, we have found that the popular choice of taking a pessimistic estimate (i.e. $\min$) is not optimal in our experiments. Instead we treat the ``aggregation function'' as a hyperparameter, and tune it over the choices of $\min, \max, \mathrm{mean}$. Across all environments, we have found $\mathrm{max}$ to be the optimal choice for aggregation.
We use two online networks and two corresponding target networks for updates, and upon calculating an aggregated estimate of $u_{\bar{\psi}}$ (shown in Eq.~\eqref{eq:u-update}) we use the $\mathrm{max}$ over the two target networks, calculated $(s,a)$-wise: $u_{\bar{\psi}}(s,a)=\max \left\{u^{(1)}_{\bar{\psi}}(s,a), u^{(2)}_{\bar{\psi}}(s,a) \right\}$. We train both of the online networks independently to minimize the same loss in Equation~\eqref{eq:u-loss}, measured against the aggregated target value. This is the same technique employed in \cite{fujimoto2018addressing} for discounted un-regularized RL and in \cite{SAC2} for discounted regularized RL.
We use the same aggregation method ($\mathrm{max}$) on the online networks $u^{(i)}_\psi$ to calculate $\theta$ across the current batch of data via Equation~\eqref{eq:theta-batch}.


\subsection{Posterior Policy Iteration\label{sec:ppi}}

Although we plot the greedy policy's reward as an evaluation metric, EVAL inherently aims to maximize the entropy-regularized reward-rate (a combination of rewards and policy relative entropy) shown in Equation~\eqref{eq:theta-defn}. Thus, the ERAR MDP's corresponding optimal policy is necessarily stochastic as discussed. The greedy policy used in evaluation, $\hat{\pi}(a|s)=\text{argmax}_a \pi_0(a|s) u(s,a)$ may therefore not be the correct policy for maximizing the un-regularized average reward rate.


Instead, it may be of interest to directly obtain the true greedy solution corresponding to the un-regularized average-reward RL objective ($\beta \to \infty$) shown in Equation~\eqref{eq:reward-rate-defn}. A simple way to find such a solution is to slowly increase the value of $\beta$ throughout training, as done in \cite{SAC1}. However, this can lead to numerical instabilities because of the exponential term $\exp \beta \left( r(s,a) - \theta \right)$ in Equation~\ref{eq:eigvec}. One workaround was proposed by an updated version of SAC \cite{SAC2}, where the authors implement a learning-based method of temperature annealing based on entropy constraints in the dual space, facilitated by Lagrange multipliers. 

\begin{algorithm}[t!]
  \caption{Posterior Policy Iteration (PPI)}
  \label{alg:rawlik}
  \begin{algorithmic}
    \STATE \textbf{Initialize}: Prior policy $\pi_0$, $\beta>0$, solve budget.
    \WHILE {$N < $ solve budget}
      \STATE $\pi_0 \gets \text{Solve}(\pi_0, \beta)$
    \ENDWHILE
    \STATE \textbf{Output}: Deterministic optimal policy $\pi^*_{\beta=\infty}=\pi_0$
  \end{algorithmic}
\end{algorithm}

\begin{figure*}[t]
\begin{minipage}[t]{1.0\textwidth}
  \includegraphics[width=\linewidth, trim={0 2cm 0 0},clip]{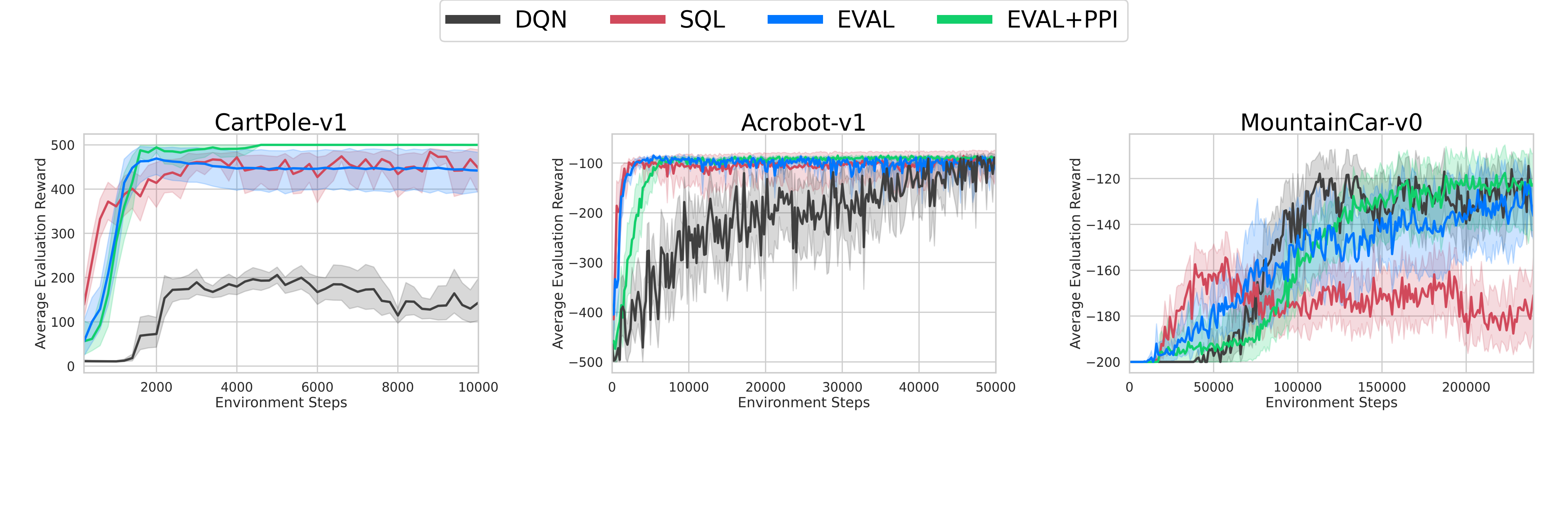}
  \caption{Classic control benchmark comparing soft Q-learning (SQL), deep Q network (DQN) and our two proposed methods (EVAL, EVAL+PPI). We find EVAL and EVAL+PPI to generally obtain higher reward with less variance than SQL or DQN.}
  \label{fig:classic-control}
\end{minipage}

\end{figure*}

A different method of recovering the un-regularized solution ($\beta \to \infty$) is based on an observation by \cite{rawlik-thesis}, which we outline below. Consider for notational convenience the function\ $\text{Solve}: (\mathbb{R}^{|\s||\A|},\Delta(\A), \mathbb{R}_+) \to \Delta(\A)$, which solves an ERAR MDP for a given choice of reward function, prior policy $\pi_0 \in \Delta(\A)$, and inverse temperature $\beta$ by outputting the corresponding optimal policy, $\pi^* \in \Delta(\A)$ (which maximizes the objective in Equation~\eqref{eq:theta-defn}). Then, Theorem 4 in \cite{rawlik2012stochastic} indicates that the iteration shown in Algorithm~\ref{alg:rawlik} will converge to the greedy optimal policy: the solution of the un-regularized MDP. Thus, we implement this ``Posterior Policy Iteration'' (PPI) technique using a function approximator (MLP) to parameterize the prior policy. Additionally, we note that this method has the benefits of not requiring a set temperature, an ``annealing schedule'' $\beta_t \to \infty$ or requiring a minimum entropy with additional computational complexity for a learning scheme \cite{SAC2}. These benefits come at the expense of some additional overhead (the memory and training of another network) and two new hyperparameters: the prior network's update frequency and update weight (Polyak parameter, $\tau_\phi, \omega_\phi$ in Algorithm~\ref{alg:u-ppi}).

To update the prior as prescribed in Algorithm~\ref{alg:rawlik}, we periodically replace the policy $\pi_0$ with the current estimate of the optimal policy $\pi^*(a|s) \propto \pi_0(a|s) u(s,a)$ at a fixed frequency throughout training. Given sufficient time for convergence at each iteration, this process is guaranteed to converge to the solution for $\beta \to \infty$, thereby solving the average-reward MDP without entropy-regularization. In practice, we do not perform hard updates of the prior network's weights, and instead employ Polyak averaging between the parameters of an online prior and a target prior. The online prior is trained to minimize the Kullback-Leibler (KL) divergence to the (current estimate of the) optimal policy via Equation~\eqref{eq:prior-loss}:
\begin{align}
    &\mathcal{L}_\phi =\frac{1}{|\mathcal{B}|}\sum_{\{s,a,r,s'\}\in \mathcal{B}} \text{KL}\left(\pi^{\bar{\phi} \psi}(\cdot |s)  \mid \pi^\phi_0(\cdot|s)\right) 
    \label{eq:prior-loss}
\end{align}

where $\mathcal{B}$ is a mini-batch randomly sampled from the replay buffer and $\pi$ is the current estimate for the optimal policy calculated from a combination of the online $u$ network and target $\pi_0$ network according to Equation~\eqref{eq:optimal-policy},

\begin{equation}
    \pi^{\bar{\phi} \psi}(a|s) = \frac{\pi^{\bar{\phi}}_0(\cdot|s)u^\psi(s,\cdot)}{\sum_a\pi^{\bar{\phi}}_0(a|s)u^\psi(s,a)}.
\end{equation}


Notice that since the current estimate of the policy is dependent upon the estimate of $\pi_0$, we use the online version of both to select actions. Further implementation details and pseudocode of this extension are provided in Algorithm \ref{alg:u-ppi} below. Experimentally, we find that even for short timescales of iteration (compared to the timescale for a full solution), the PPI method is still able to recover the $\beta \to \infty$ solution, in agreement with Section 4.1.2 of \cite{rawlik-thesis}.

\section{Experiments \label{sec:experiments}}
We focus on the classic control environments from OpenAI's Gymnasium \cite{brockman2016openai} \nocite{pmlr-v139-ceron21a} with varying levels of complexity: CartPole-v1, Acrobot-v1, MountainCar-v0. First we will provide benchmark experiments, comparing EVAL against DQN \cite{nature-dqn} (as it shares many implementation details) and SQL \cite{HaarnojaSQL} (for its entropy-regularized objective). Although we recognize that these algorithms are not designed to optimize the same objective function, we do not have other benchmarks with which to compare in the discrete-action setting. On the other hand, we emphasize that the metric of interest in discounted RL is often the average evalulation reward. Against this metric, EVAL and EVAL-PPI have stronger performance in terms of sample complexity compared to DQN and SQL. 

We compare EVAL with fixed (tuned) $\beta$ and with PPI (denoted EVAL and EVAL-PPI, respectively) to DQN \cite{stable-baselines3} and SQL \cite{HaarnojaSQL} on classic control environments in Figure ~\ref{fig:classic-control}. Hyperparameter tuning has been performed for each algorithm on each environment, with the associated values shown in Appendix~\ref{app:hparams}.

For proper comparison with other un-regularized methods, the evaluation phase for EVAL uses the greedy policy derived from the learned stochastic policy. Every one thousand environment steps, we pause the training to perform a greedy evaluation, averaged over ten episodes. The resulting reward from each algorithm is averaged over twenty random initializations. All code for reproducing the experiments is available at \JA{\url{https://anonymous.4open.science/r/EVAL-D25E}}.

\begin{figure}[t]
    \centering
    \includegraphics[width=0.4\textwidth]{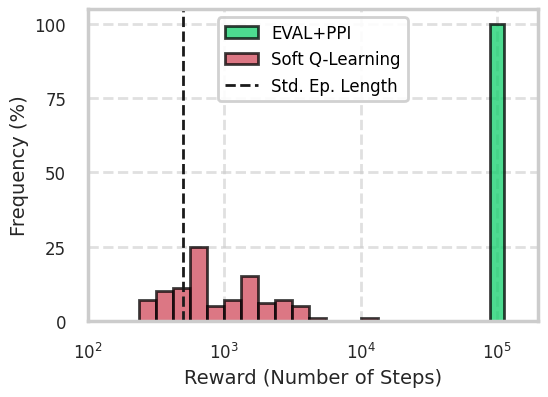}
    \caption{As a demonstration of the usefulness of EVAL+PPI, we consider a modified version of CartPole which represents a continuing task. After training for 5000 steps (in the standard CartPole-v1 environment with a maximum episode length of 500), we compare the evaluation performance of SQL with EVAL+PPI. Specifically, we set the time-limit of the environment much higher: to 100,000 steps. We find that EVAL+PPI consistently reaches the maximum number of steps while SQL only rarely achieves similarly high reward. We find that EVAL+PPI can continue episodes for at least 10~billion steps (as of submission).}
    \label{fig:continuing-task}
\end{figure}

Since the average-reward objective is aimed at solving continuing tasks, we devise a paradigmatic experiment (inspired by Fig. 2 in ~\cite{ARTRPO}) based on CartPole. showcasing the ability of EVAL in the long-time setting. Although both agents are trained in environments with a default (500) timestep limit, the EVAL agent is able to generalize further. The results of this experiment, shown in Figure~\ref{fig:continuing-task}, indicate that EVAL+PPI is far superior to SQL at solving the (effectively) continuing task of balancing the pole forever. 

\section{Limitations and Future Work \label{sec:future-work}}
The current work focuses on the case of deterministic transition dynamics. \cite{UAI} has developed a method for the more general case of stochastic transition dynamics, by iteratively learning biases for the dynamics and rewards. With a model-based algorithm, this seems to be a promising avenue for future exploration for the general case of stochastic transition dynamics. While we have focused on the left eigenvector of $\widetilde{P}$, the right eigenvector $v(s,a)$ can also be learned online. Learning $v$ gives access to the optimal steady-state distribution ($u v$) a quantity of interest as mentioned in recent work \cite{saxena2023off}. Although we have tested our method on classic control environments, other more challenging benchmarks such as the ALE \cite{ALE} can be used as well. Preliminary experiments show that EVAL is able to find the optimal policy in some Atari environments, but further hyperparameter tuning is needed to improve its stability and sample efficiency.

Our current method features discrete action spaces, but adapting the present algorithm to actor-based approaches could allow for a straightforward extension to continuous actions, analogous to the soft actor-critic approach~\cite{SAC1}. Since we use an off-policy method, it would be interesting to apply specific exploration policies for rollout collection. Since PPI \cite{rawlik2012stochastic} does not require a particular intitial policy, this method can similarly benefit from an improved exploration policy at initialization. As a value-based technique, other ideas from the literature such as $\text{TD}(n)$, REDQ \cite{redq}, PER \cite{schaul2015prioritized}, or dueling architectures \cite{wang2016dueling} may be included. An important contribution for future work is studying the sample complexity and convergence properties of the proposed algorithm. Further work may also provide a connection between PPI and temperature annealing procedures, and port this idea to the discounted framework as well.

\section{Conclusion}
By leveraging the connection of the ERAR objective to the solution of a particular eigenvalue problem, we have presented the first solution to deterministic ERAR MDPs in continuous state spaces by use of function approximation. Our experiments suggest that EVAL compares favorably in several respects to DQN and Soft Q-Learning. Our algorithm leverages the existing DQN framework allowing for a straightforward and easily extendable model applicable to the ERAR objective. We have also provided a solution to the un-regularized, greedy average-reward objective through an iterative procedure in the prior policy space. Our algorithms show an advantage in the classic control suite, especially given that there are no other algorithms for this setting. We believe that the average-reward objective with entropy regularization is a fruitful direction for further research and real-world application, with this work addressing an important gap in the existing literature.

\section{Acknowledgements}
JA would like to acknowledge the use of the supercomputing facilities managed by the Research Computing Department at UMass Boston; the Unity high-performance computing cluster; and funding support from the Alliance Innovation Lab – Silicon Valley. RVK and JA would like to acknowledge funding support from the NSF through Award No. DMS-1854350 and PHY-2425180. ST was supported in part by NSF Award (2246221), PAZY grant (195-2020), and WCoE, {Texas~Tech~U.} This work is supported by the National Science Foundation under Cooperative Agreement PHY-2019786 (The NSF AI Institute for Artificial Intelligence and Fundamental Interactions, http://iaifi.org/).


\bibliography{aaai25}
\newpage
\onecolumn
\setcounter{section}{0}
\section{Theory \label{app:theory}}




For convenience, we provide the update equations from \cite{PRR} (Eq. 26, 27) which will be learned with EVAL:
\begin{align}
    u(s,a) &\xrightarrow{} (1-\alpha)u(s,a) + \alpha e^{\beta (r(s,a)-\theta)} u(s',a') \label{eq:prr-u} \\
    e^{\beta \theta} &\xrightarrow{} (1-\alpha_\theta)e^{\beta \theta} + \alpha_\theta e^{\beta r(s,a)} \frac{u(s',a')}{u(s,a)}
\end{align}
which resembles our Eq.~\eqref{eq:u-update} and \eqref{eq:theta-batch}, respectively.

An important distinction between Eq. 26 in \cite{PRR} and Eq.~\eqref{eq:prr-u} is that we have changed the sign of $\theta$ in their definition. This allows us to write the original objective in Eq.~\eqref{eq:theta-defn} in a way consistent with the average-reward literature (e.g. Eq.~\eqref{eq:reward-rate-defn}).

\section{Experimental Details \label{app:experiments}}

\subsection{Implementation\label{app:implementation}}

\begin{algorithm}
  \caption{EVAL with PPI}
  \label{alg:u-ppi}
  \begin{algorithmic}
    \STATE \textbf{IN:} sample budget, environment, $\beta$, hyperparameters
    \STATE \textbf{Initialize}:
        \STATE Online network weights: $\psi_i, {\color{blue}\phi_i} \sim W(\cdot)$
        \STATE Target network weights: $\bar{\psi}_i = \psi_i, \color{blue}\bar{\phi_i} = \phi_i$
        \STATE Reward-rate: $\theta = 0$
        \STATE Replay buffer: $\mathcal{D} = \{\}$
        
    \WHILE {$t < $ sample budget}
        \STATE Collect experience:
            \STATE Sample action $a\  {\color{blue} \sim \pi^\phi_0 \propto u^\psi \cdot \pi_0^{\widehat{\phi}}}$
            \STATE Take step in environment $s' \sim p(\cdot | s,a)$
            \STATE Save to replay buffer: $\mathcal{D} \leftarrow \{s, a, r, s'\}$
            \FOR{each gradient step}
                \STATE Sample minibatch  $\mathcal{B} \subset \mathcal{D}$ 
                \STATE Calculate loss via Eq.~\eqref{eq:u-loss} and Eq.~\eqref{eq:prior-loss}
                \STATE Do gradient descent using Adam on $u$ networks {\color{blue} and $\pi_0$ network}
                \STATE {\color{blue}Calculate $\theta_\text{new}$ via  Equation~\eqref{eq:theta-batch}} 
                \STATE Update $\theta$ estimate: $\theta \gets \theta \cdot(1-\tau_\theta) + \theta_{\text{new}}\cdot \tau_\theta$
            \ENDFOR
        \IF{update target params}
        \STATE Update target parameters (Polyak averaging with parameters $\tau_\psi, {\color{blue}\tau_\phi}$).
        \ENDIF

    \ENDWHILE
    \STATE \textbf{OUT:} Optimal policy for AR-MDP
  \end{algorithmic}
\end{algorithm}
We implement our algorithm in PyTorch, and follow the style of Stable Baselines3 \cite{stable-baselines3}. 
The function approximators for $u(s,a)$ (and similarly for $\pi_0$, discussed below) are MLPs with state as input, and $|\A|$ output heads. 
Below, we show the pseudocode for EVAL with the PPI method (a combination of Algorithm~\ref{alg:u} and \ref{alg:rawlik} in the main text), allowing one to solve the un-regularized RL objective. We highlight the differences from Algorithm \ref{alg:u} (main text) in {\color{blue}blue}. As mentioned in the main text, we require a parameterization for the (online and target) prior policy as well as an update frequency and rolling average weight for the Polyak update.

We highlight the use of Eq.~\eqref{eq:theta-batch} in this algorithm, as it involves an expectation operation taken over the prior policy. We take the target prior for this calculation, as opposed to the online prior network for the calculation of the optimal policy, as seen in the loss function of Eq.~\eqref{eq:prior-loss}.

\subsection{Choice of Architecture}
We note that since $u(s,a) > 0$ (by the Perron-Frobenius theorem), we require the $u$ network outputs to be strictly positive and thus use a soft-plus activation function at the output layer. Although a ReLU activation is possible, we found it to give much worse performance across all tasks, despite additional hyperparameter tuning.

We similarly parameterize the prior net as an MLP (same hidden dimension as $u$-network), but with a softmax output. We have experimented with sharing weights between $u$ and $\pi_0$ networks without finding significant performance gains, but this may be an interesting avenue for future exploration.

\subsection{Soft Q-Learning}
We implement our own soft Q-learning algorithm, in a similar style as  EVAL (i.e. with a minimization over two networks, and use of a separate lagging target network). Similar to EVAL's evaluation, we take the greedy policy $\pi(s)=\textrm{argmax}_a Q(s,a)$ for evaluating the agent.

\subsection{Hyperparameters\label{app:hparams}}
\begin{center}
\captionsetup{type=figure}
\captionof{table}{Hyperparameter Ranges Used for Finetuning EVAL \& PPI}
\begin{tabular}{||c c||} 
 \hline
 Hyperparameter & Value Range\\ [1ex] 
  \hline
 Learning Rate, $\eta$ & $(10^{-4} , 10^{-1})$ \\ 
 \hline
 Inverse Temperature, $\beta$ & $0.01, 0.05, 0.1, 0.5, 1, 2, 5, 10$ \\
 \hline
 Batch Size, $b$  & $16, 32, 64, 128, 256$ \\ 
 \hline
 Target Update Interval, $\omega$ &  $10, 100, 500, 1000, 2000, 5000$ \\ [1ex] 
 \hline
 Gradient Steps, UTD & 1,5,10,50 \\ [1ex] 
 \hline
 
\end{tabular}
\label{tab:sweep}
\end{center}
For each run, we choose the buffer size to be the same number of steps in the sample budget (except for DQN, which has a tuned value of buffer size $B$, given by \cite{stable-baselines3}). We also tune the hidden dimension ``hid. dim.'' of the MLP over values in $\{ 32, 64, 128\}$. 
In the following Table~\ref{tab:hparams-u}, $\omega_\pi$  refers to the prior policy's target network update interval (in terms of environment steps).

For simplicity, we have found the algorithms to generally work well for some fixed hyperparameters, reducing the search space and potential sensitivity: The Polyak update ratio (for $u$ and $\pi_0$ in PPI) is fixed to 1.0 (``hard updates''); The ERAR rate, $\theta$ is kept frozen at zero (except for irreducible tabular dynamics); the number of gradient steps per environment step is fixed to $5$ throughout; The replay buffer size set to the maximum value (total number of environment steps during training); and the agent is trained after every environment step (cf. ``training frequency'' in \cite{stable-baselines3}). The finetuned hyperparameters for each environment are listed below. Each is the result of a sweep of roughly 200 runs (in random search), each:

\begin{center}\captionsetup{type=figure}
\captionof{table}{Finetuned Hyperparameter Values for EVAL(-PPI)}
\begin{tabular}{||c c c c c c c c||} 
 \hline
 Environment & hid. dim. & $\eta$ & $b$ & $\beta$ & $\omega$ & learn starts & $\JA{\omega_\pi}$ \\ [0.5ex] 
  \hline
 CartPole-v1 & 16 & $1 \cdot 10^{-3}$ & $64$ & 2 & 10 & 0  & 500\\ [0.5ex] 
 \hline
 Acrobot-v1 & 64 & $5 \cdot 10^{-4}$ & 64 & 0.01 & 10 & 0  & 500 \\ [0.5ex] 
 \hline
 MountainCar-v0 & 32 & $6 \cdot 10^{-4}$ & 128 & 20 & 100 & 5000  & 2000 \\ [1ex] 
 \hline
\end{tabular}
\label{tab:hparams-u}
\caption{ The final two columns show the additionally tuned (with all others held fixed) hyperparameters specific to PPI.}
\end{center}

\begin{center}\captionsetup{type=figure}
\captionof{table}{Finetuned Hyperparameter Values for DQN}
\begin{tabular}{||c c c c c c c c c c c ||} 
 \hline
 Environment & $\eta$ & $b$ & $B$ & $\gamma$ & $\omega$ &  $\varepsilon_{\textrm{final}}$ & $\varepsilon_{\textrm{frac}}$ & grad step & train freq & learn starts\\ [0.5ex] 
  \hline
 CartPole-v1 & $2.3 \cdot 10^{-3}$ & 64 & 100,000 & 0.99 & 10 & 0.04 & 0.16 & 128 & 256 & 1000 \\ 
 \hline
 Acrobot-v1  & $6.3 \cdot 10^{-4}$ & 128 & 50,000 & 0.99 & 250 & 0.1 & 0.12 & -1 & 4 & 0\\ 
 \hline
 MountainCar-v0  & $4 \cdot 10^{-3}$ & 128 & 10,000 & 0.98 & 600 & 0.07 & 0.2 & 8 & 16 & 1000 \\  [1ex] 
 \hline
\end{tabular}
\label{tab:hparams-dqn}
\caption{ $\varepsilon_{\textrm{frac}}$ denotes the exploration fraction over which to decay $\varepsilon=1.0$ to $\varepsilon=\varepsilon_{\textrm{final}}$. A training frequency of $-1$ indicates that training of the $Q$ networks occurs only after the end of each rollout episode. The provided optimal values for $\tau=1.0$ and hidden dimension of 256 throughout all environments.}
\end{center}

\begin{center}\captionsetup{type=figure}
\captionof{table}{Finetuned Hyperparameter Values for SQL}
\begin{tabular}{||c c c c c c c c c c||} 
 \hline
 Environment & hid. dim & $\eta$ & $b$ & $\beta$ & $\gamma$ & $\tau$ & $\omega$ & grad steps & learn starts \\ [0.5ex] 
  \hline
 CartPole-v1 & 64 & $2 \cdot 10^{-2} $ & 64 & 0.1 & 0.98 & 0.95 & 100 & 9  & 1,000\\ [1ex] 
 \hline
 Acrobot-v1 & 32 & $6.6 \cdot 10^{-3} $ & 128 & 2.6 & 0.999 & 0.92 & 100 & 9 & 2,000 \\ [1ex] 
 \hline
 MountainCar-v0 & 64 & $2\cdot 10^{-3} $ & 128 & 0.7 & 0.99 & 0.97 & 100 & 2 & 9,000\\ [1ex]
 \hline
\end{tabular}
\label{tab:hparams-sql}
\caption{We use the finetuned hyperparameters given in \cite{adamczyk2024boosting} for soft $Q$-learning in the classic control benchmark.}
\end{center}



We use implementations of DQN from stable-baselines3, with the finetuned hyperparameters for each environment given by \cite{stable-baselines3}.

\section{Additional Results}

\subsection{Number of networks}
\begin{figure}[h]
    \centering
    \includegraphics[width=0.5\textwidth]{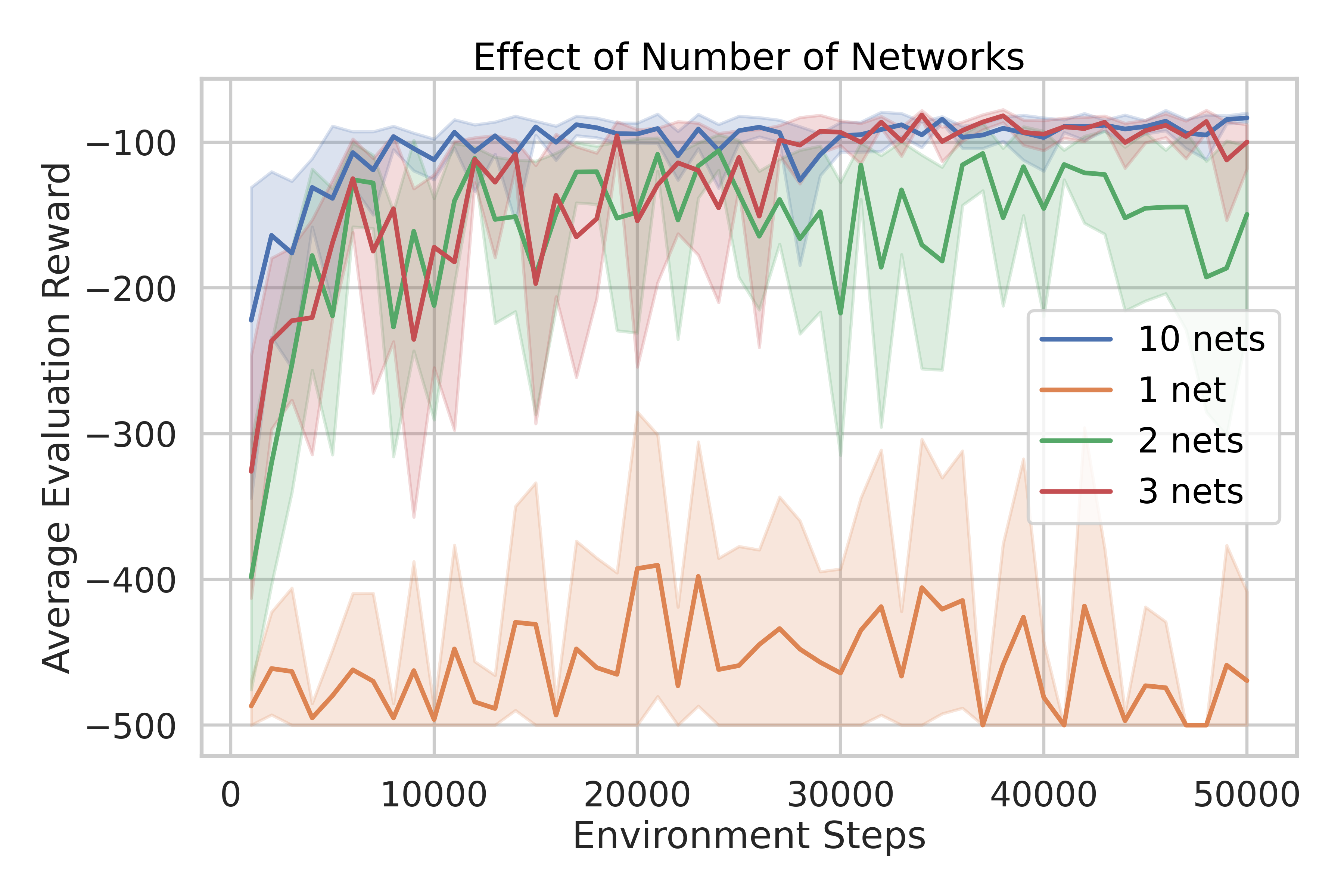}
    \caption{For EVAL (without PPI) we compare the performance for training multiple networks in parallel. All networks are aggregated with the $\mathrm{max}$ function as discussed in the ``Proposed Algorithms'' section.}
    \label{fig:ablation-num-nets}
\end{figure}

In light of several recent works maintaining multiple estimates or an entire ensemble of the agent's $Q$ function \cite{sunrise, redq}, we consider the effect of learning more than two (online and target) networks on training performance. The results shown in Figure~\ref{fig:ablation-num-nets} indicate that having greater than one or two networks can drastically improve performance in some environments, with diminishing returns as seen in prior work. Our experiments indicate that a minimum of two networks is required for learning the optimal policy, so we use two networks throughout for simplicity, despite this generally being a tunable parameter. For convenience, we use two independent target networks as well, to separately perform rolling averages of the target parameters before aggregating.
\end{document}